# Antagonistically Actuated Rolling Contact Joint Mechanism

Shuya Wada, Shota Miyake, Kazuki Abe, Masahiro Watanabe, and Kenjiro Tadakuma

*Abstract*— **Rolling contact joints such as the Hillberry joint offer low friction, low wear, and a wide range of motion, but their moving center of rotation generally requires transmission elements such as tendons, which introduce friction, wear, and routing constraints. This paper proposes a soft active Hillberry joint mechanism in which soft pneumatic bending actuators placed at the rolling contact drive the joint directly, and their antagonistic arrangement enables bidirectional rotation while preserving rolling contact. To make the joint angle easier to adjust, we aimed to bend the actuators sequentially from the air supply side by narrowing the flow channel to delay pressure propagation, and compared two actuators with different flow-channel cross-sectional areas. Narrowing the channel increased the force required for bending by approximately 0.1–0.2 N and did not produce the intended pressure propagation delay, likely because differences in actuator expansion outweighed the effect of flow restriction. A prototype using the actuator with the wider channel had a joint angle range limited to −92.5° to 49.0° at 60 kPa, and its joint torque exhibited hysteresis that increased with supply pressure. These results suggest that the deformation resistance of the antagonistic actuator and the wires constraining the links, which restrict actuator expansion, are the main limiting factors. Based on these findings, we present design guidelines for soft-actuated rolling contact joints: reducing the deformation resistance of the antagonistic actuator and imparting anisotropic deformation to the actuators themselves.**

## I. INTRODUCTION

Robotic joints are required to achieve both a wide range of motion and low friction and wear. A joint supports loads while allowing relative motion between links, and thus determines the range of motion, stiffness, load capacity, and safety during contact with the environment. Conventional revolute joints constrain the center of rotation with a shaft and bearings, but their sliding parts and clearances lead to friction, wear, backlash, and an increased number of components. Compliant mechanisms that exploit the elastic deformation of members have been studied to avoid these issues [1], [2]; however, their range of motion is narrow because the motion is limited to the range of elastic deformation. In contrast, rolling contact joints, which generate relative motion through rolling between contact surfaces, achieve large rotation without sliding.

The Hillberry joint, a type of rolling contact joint, consists of two cylindrical surfaces in rolling contact without slipping, constrained to each other by crossed straps wrapped around them [3]. It exhibits low friction and wear, can bear compressive loads at the contact surfaces, and provides a wide range of motion because both links rotate. Owing to these features, applications such as artificial finger joints [4], the ARC joint with kinematically variable torsional stiffness [5], mechanisms whose transmission characteristics are programmed by noncircular contact surfaces [6], and joints with a wide range of motion for inflatable robots [7] have been reported.

The authors are with the Graduate School of Engineering Science, The University of Osaka, Toyonaka, Osaka 560-8531, Japan (e-mail: kenjiro.tadakuma.es@osaka-u.ac.jp).

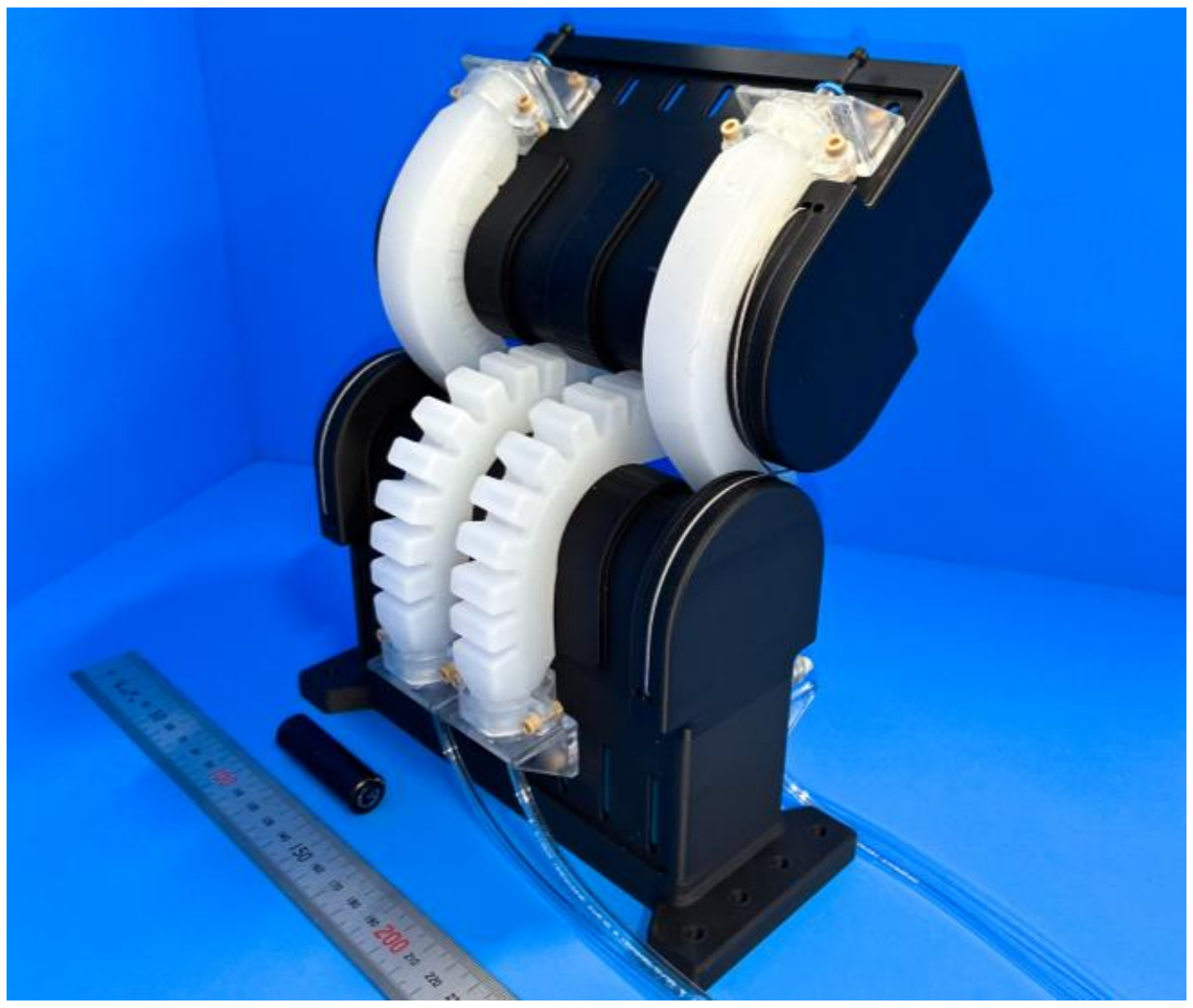

Fig. 1. Prototype of the soft active Hillberry joint mechanism, in which the rolling contact section is driven by antagonistically arranged soft pneumatic bending actuators.

However, because the center of rotation of a rolling contact joint moves with the contact point and the joint has no fixed rotation axis, it is difficult to connect a rotary actuator directly to the joint. Therefore, in most robotic applications, the rolling contact is driven by pulling tendons routed through the links [8], [9]. Although tendon drive allows the actuators to be located away from the joint, it entails three issues arising from the transmission elements between the joint structure and the actuators. First, friction between the tendons and pulleys or sheaths impairs force transmission, causing position transmission errors and hysteresis [10]–[12]. Second, repeated bending and sliding cause tendon wear and fatigue, requiring pretension management to compensate for the tension loss due to elongation [13]. Third, the tendon routing is constrained by the link geometry and joint arrangement, requiring designs that account for path length changes with joint angle and coupling between joints [14].

One way to eliminate transmission elements is to give the joint structure itself the function of an actuator. Soft pneumatic actuators have no sliding parts and can therefore be easily integrated with structural members, and they can generate bending deformation by pressurization alone [15]. For example, pneumatic network (PneuNet) actuators, in which multiple chambers inside an elastomer are connected by a flow channel, achieve large bending deformation with a simple structure [16], [17].

The pressure propagation delay in soft pneumatic actuators has been exploited to sequence motions, but not to tune the motion characteristics of joints. Owing to air compressibility and the flow resistance of the channel, pressure propagation to chambers far from the supply port is delayed [18]. Although this delay is generally regarded as degrading responsiveness, methods that actively exploit it to generate sequential motions or complex deformations from a single pressure input have been proposed [19]–[22]. However, these methods mainly target motion sequencing in grippers, mobile robots, and wearable devices. Although the bending progression of a joint actuator directly determines the change in joint angle, to the best of our knowledge, no study has designed the internal channel cross-sectional area of an actuator to tune the motion characteristics of a joint.

In this study, we propose a soft active Hillberry joint mechanism in which the rolling contact section of a Hillberry joint is replaced with soft pneumatic bending actuators that drive the joint directly with air pressure (Fig. 1). The proposed mechanism retains the low friction and low wear of rolling contact while achieving bidirectional actuation through an antagonistic arrangement of the actuators. The contributions of this paper are threefold. First, as a new configuration for driving rolling contact joints without transmission elements, we present the principle of this mechanism and realize it in a prototype. Second, by comparing two actuators with different flow-channel cross-sectional areas, we experimentally clarify the effects of the channel area on the force required for bending and the pressure propagation delay. Third, by evaluating the range of motion and joint torque of the prototype, we identify the deformation resistance of the antagonistic actuator and the wires constraining the links as the likely main factors limiting performance, and present design guidelines for improvement.

## II. Proposed Mechanism

### A. Concept of the Soft Active Hillberry Joint

The soft active Hillberry joint mechanism retains the rolling contact structure of the Hillberry joint mechanism while using soft actuators to drive it. In conventional Hillberry joint mechanisms, joint motion is generated by externally driving the rolling contact section constrained by straps. In contrast, in the proposed mechanism, soft pneumatic bending actuators are placed at the rolling contact section and drive the joint directly with air pressure. Fig. 2 shows a schematic of the proposed mechanism. Note that the mechanism is not limited to actuators of a specific structure and can also be applied to other soft actuators capable of driving the rolling contact section by bending.

### B. Bidirectional Actuation by Antagonistic Arrangement

A single soft pneumatic bending actuator can generally generate active bending in only one direction. Therefore, in the proposed mechanism, the actuators are arranged antagonistically on the cylindrical contact surfaces of the links, as shown in Fig. 2. These actuators generate torques in opposite directions about the joint.

Fig. 3 shows the operation of the proposed mechanism. Here, the relative angle between the two links is defined as the joint angle θ, with θ = 0° when the two links are aligned in a straight line. When pressure is applied to one actuator, the actuator bends, and the links accordingly rotate relative to each other while maintaining rolling contact. This direction of rotation is defined as the positive θ direction. When the pressurized actuator is switched, the joint rotates in the opposite direction, and θ becomes negative. Thus, bidirectional active joint actuation by air pressure is achieved while retaining the low friction and low wear of rolling contact.

Furthermore, by controlling the pressure applied to each actuator individually, θ can, in principle, be held at a desired value by adjusting the balance between the torques generated by the two actuators. That is, individual pressure control of the antagonistically arranged actuators is expected to enable control of both the direction of rotation and the joint angle.

### C. Flow-Channel Design for Sequential Bending

The proposed mechanism is intended to make the joint angle easier to adjust by bending the actuator sequentially from the supply side. The soft pneumatic bending actuator used in this study consists of multiple chambers inside an elastomer connected by a flow channel, and it bends almost uniformly when pressurized. If such behavior occurs in the proposed mechanism, the joint rotates all at once, making it difficult to position the joint at a desired angle. In contrast, if bending progresses sequentially from the chamber on the supply side, the rotation at the rolling contact also progresses gradually, making the joint angle easier to adjust.

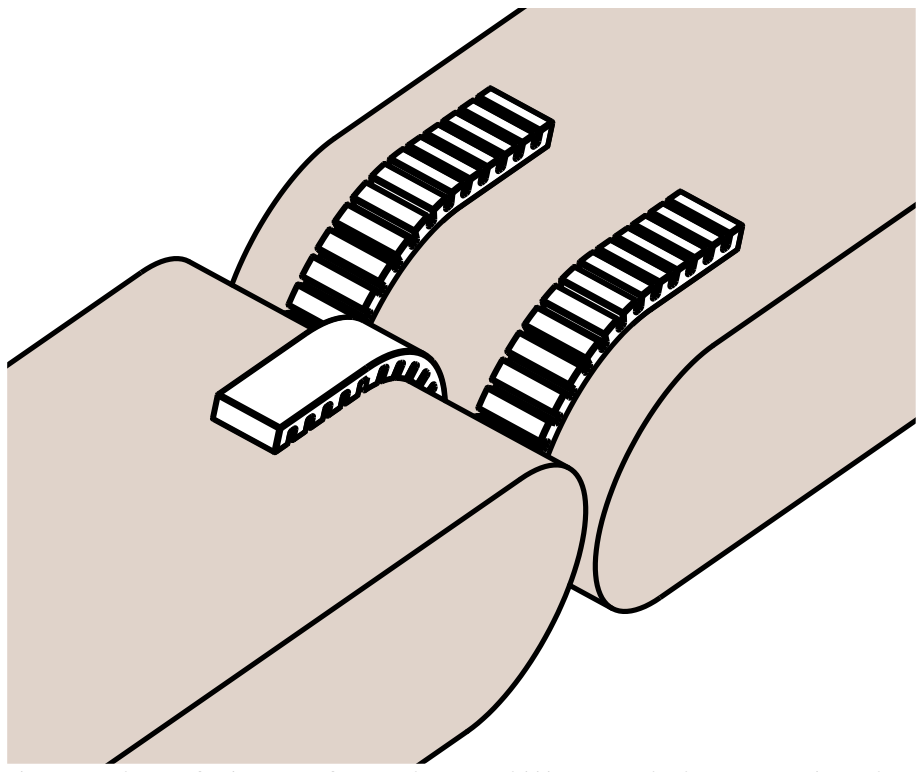

Fig. 2. Schematic of the soft active Hillberry joint mechanism. Two soft pneumatic bending actuators are arranged antagonistically at the rolling contact between two links with cylindrical contact surfaces.

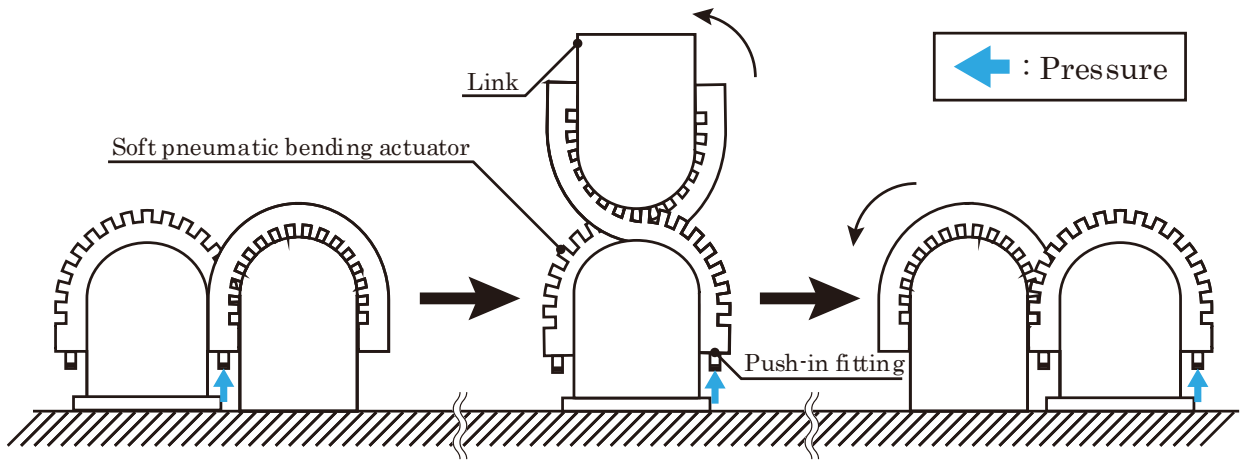

(a) Single-side pressurization

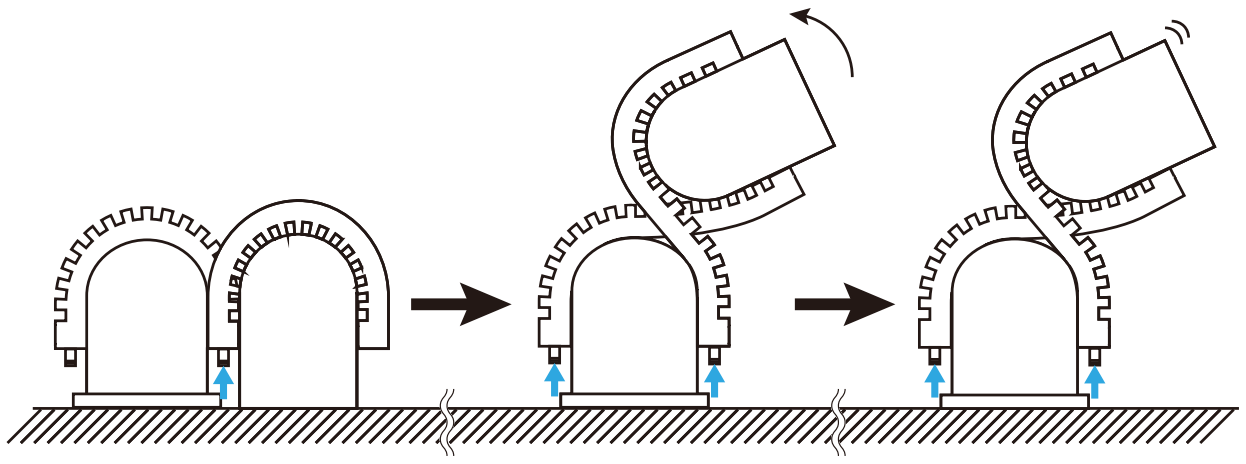

(b) Both-side pressurization

Fig. 3. Operation of the soft active Hillberry joint mechanism. (a) Single-side pressurization: pressurizing one actuator bends it and rotates the joint while the links maintain rolling contact, and switching the pressurized actuator rotates the joint in the opposite direction. (b) Both-side pressurization: the joint angle is adjusted by the balance of the pressures applied to the two actuators.

To achieve this sequential bending, two actuators with different flow-channel cross-sectional areas were designed (Fig. 4). Reducing the channel cross-sectional area increases the flow resistance of the channel and delays pressure propagation to the chambers far from the supply side, which is expected to make the actuator bend sequentially from the supply side. The two actuators have identical outer dimensions and differ only in the channel cross-sectional area. Hereafter, the actuator with the relatively larger channel cross-sectional area is referred to as the wide-channel actuator, and the one with the smaller area as the narrow-channel actuator.

### *D. Prototype Design and Fabrication*

Because a high pressure must be applied to obtain the torque required to drive the joint, silicone rubber with a Shore hardness of 20A (Dragon Skin 20, Smooth-On) was used as the base material of the actuators to withstand high pressures. The links were fabricated from PLA. The connectors between the actuators and the links were fabricated from Clear Resin V5 using a stereolithography 3D printer (Form 4, Formlabs).

To prevent the distance between the links from changing due to the elasticity and expansion of the silicone rubber, wires were attached near the contact surfaces of the links. These wires also suppress extension of the actuators in directions other than bending during pressurization. In addition, when one actuator was placed on each side, the mechanism twisted during pressurization. Therefore, two actuators were placed on each side of the prototype, for a total of four actuators.

Fig. 1 shows the prototype realized based on the above principle, and Table I lists its specifications.

## III. Actuator Characterization

This section compares the characteristics of the wide-channel and narrow-channel actuators to investigate the effects of the flow-channel cross-sectional area on the force required for bending and on the pressure propagation delay. Section III-A measures the force required for bending, and Section III-B measures the pressure propagation delay between the proximal and distal ends. Section III-C discusses the effects of the channel area based on these results and selects the actuator to be used in the joint mechanism.

Elastomers such as silicone rubber, the base material of the actuators, exhibit the Mullins effect, in which their mechanical properties change depending on the deformation history, with pronounced stress softening particularly during the first deformation [23]. Therefore, before the measurements, 60 kPa, the maximum pressure used in the experiments, was applied three times to every actuator to stabilize its properties.

### *A. Force Required for Bending*

To investigate the effect of the flow-channel cross-sectional area on the force required for bending, the relationship between displacement and force was measured with a tensile testing machine (STB-1225S, A&D), as shown in Fig. 5. One end of the actuator was fixed, and the other end was pulled from its initial position perpendicular to the longitudinal axis at 360 mm/min to bend the actuator, while the required force was measured. To evaluate the force required to bend the actuator itself, the measurements were performed without pressurization. The displacement was increased until the actuator began to extend in the longitudinal direction. To account for individual differences between actuators and variation between measurements, three actuators of each type were used, and each actuator was measured five times.

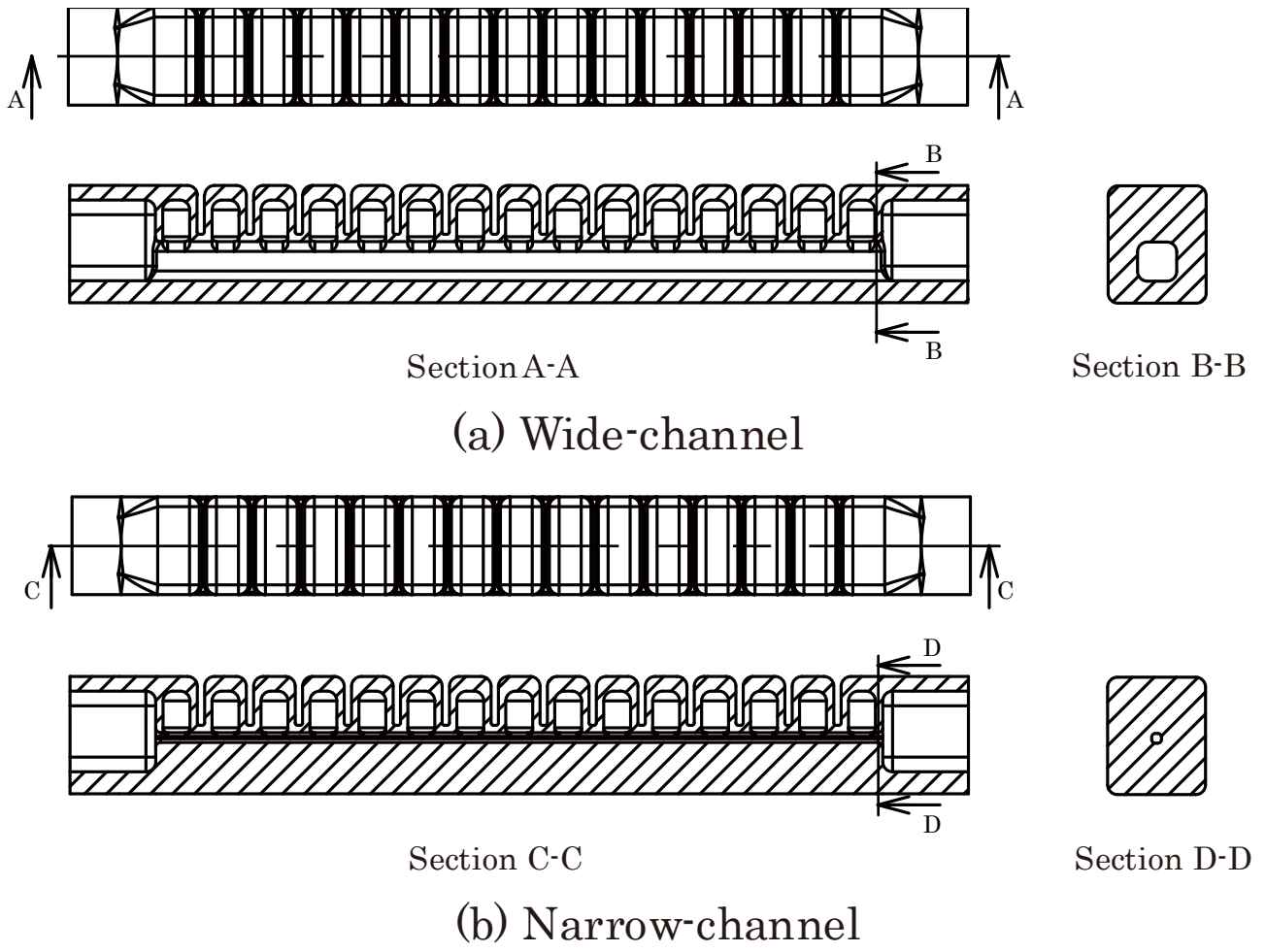

Fig. 4. Cross-sectional views of the two types of soft pneumatic bending actuators: (a) wide-channel actuator and (b) narrow-channel actuator. The two actuators have the same outer dimensions and differ in the cross-sectional area of the flow channel.

TABLE I. Specifications of the Prototype

| Parameter | Wide-channel | Narrow-channel |
|---|---|---|
| Mechanism dimensions [mm] (W × D × H) | 165 × 98 × 250 | 165 × 98 × 250 |
| Mechanism mass [g] | 670.0 | 698.6 |
| Actuator dimensions [mm] (W × D × H) | 230 × 38 × 27 | 230 × 38 × 27 |
| Actuator mass [g] | 98.6 | 105.8 |
| Cross-sectional area [mm²] | 60.6 | 3.5 |

The narrow-channel actuator required more force for bending than the wide-channel actuator over the entire displacement range (Fig. 6). The difference was approximately 0.12–0.15 N up to a displacement of about 110 mm and increased to 0.23 N near the maximum displacement, where the force was 1.23 ± 0.05 N for the narrow-channel actuator and 1.00 ± 0.02 N for the wide-channel actuator (mean ± SD, n = 15). For both actuators, the required force increased nonlinearly with displacement.

### *B. Pressure Propagation Delay*

To investigate the effect of the flow-channel cross-sectional area on the pressure propagation delay, the pressures at the proximal and distal ends of the actuator were measured using pressure sensors (DP-102A-E-P, Panasonic) and an oscilloscope (MSO46, Tektronix). The experimental setup is shown in Fig. 7. With the actuator suspended, pressure was supplied from the proximal end, and the pressures at the proximal and distal ends were measured simultaneously. The supply pressure was varied from 10 to 50 kPa in steps of 10 kPa, and three measurements were performed for each condition.

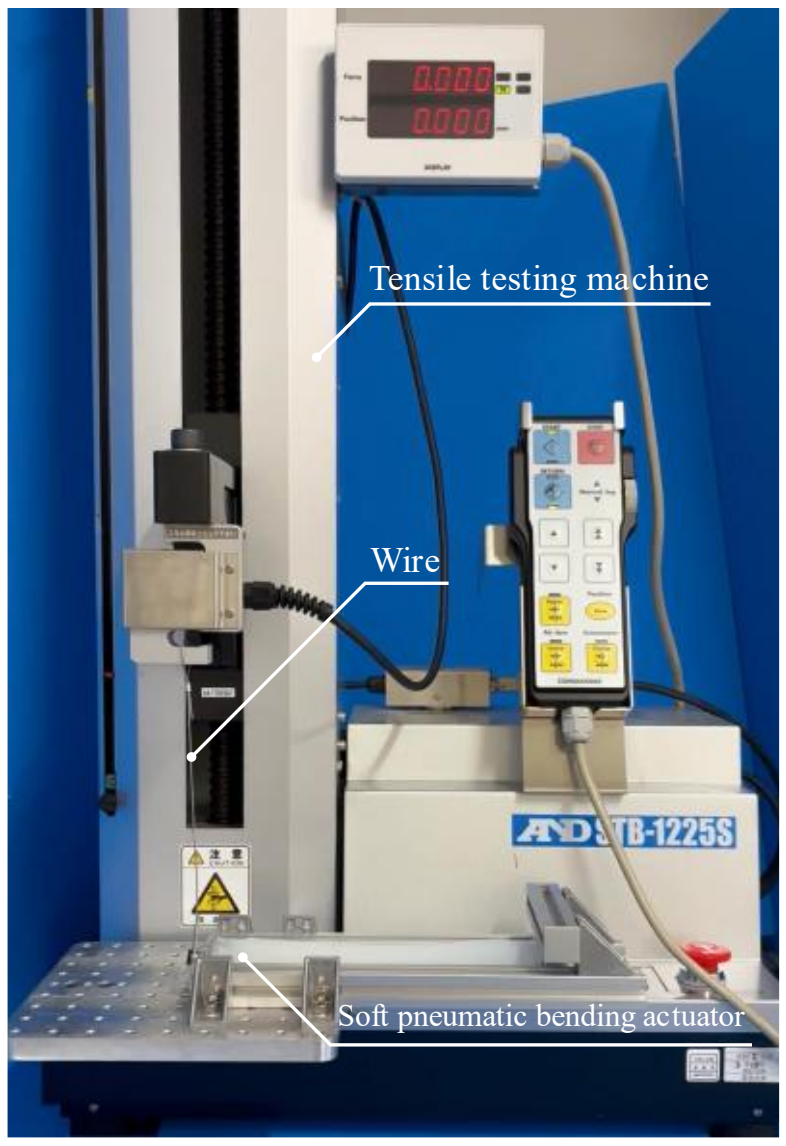

Fig. 5. Experimental setup for measuring the force required for bending.

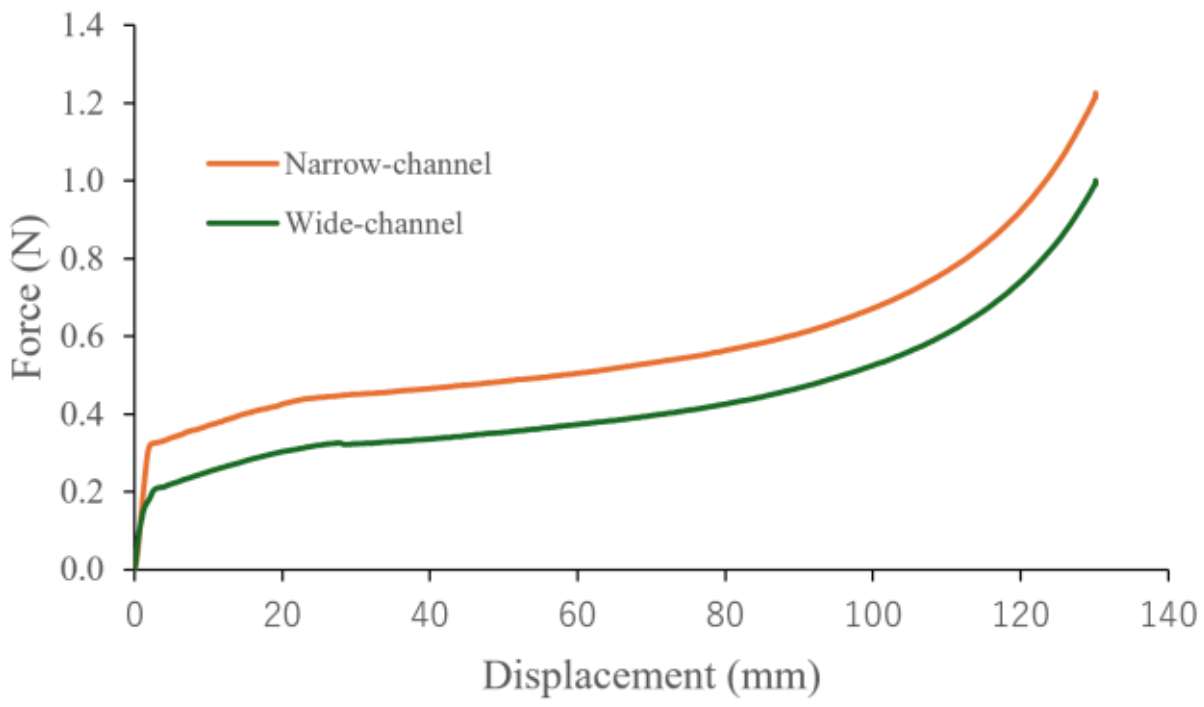

Fig. 6. Force required for bending versus displacement for the wide-channel and narrow-channel actuators (mean, n = 15).

The evaluation was based on the times at which each sensor output reached 10% and 90% of its change from the initial value to the steady-state value, hereafter referred to as the 10% and 90% response times. The 10% response time served as an index of the onset of the response, and the 90% response time as an index of its completion. Using these two times reduces the influence of initial noise and of the asymptotic behavior near the steady-state value. For each of these times, the difference between the distal and proximal ends is defined as the response time difference.

The response time difference of the narrow-channel actuator clearly exceeded that of the wide-channel actuator only for the 10% response time at a supply pressure of 10 kPa (Fig. 8). Under this condition, the response time difference of the narrow-channel actuator was approximately 40 ms larger, whereas at 20 kPa or higher, the difference between the two actuators was 2 ms or less. In contrast, for the 90% response time, the response time difference of the wide-channel actuator was larger than that of the narrow-channel actuator at all supply pressures, and the difference increased with supply pressure. At 50 kPa, the response time difference was 190.0 ms for the wide-channel actuator and 85.3 ms for the narrow-channel actuator.

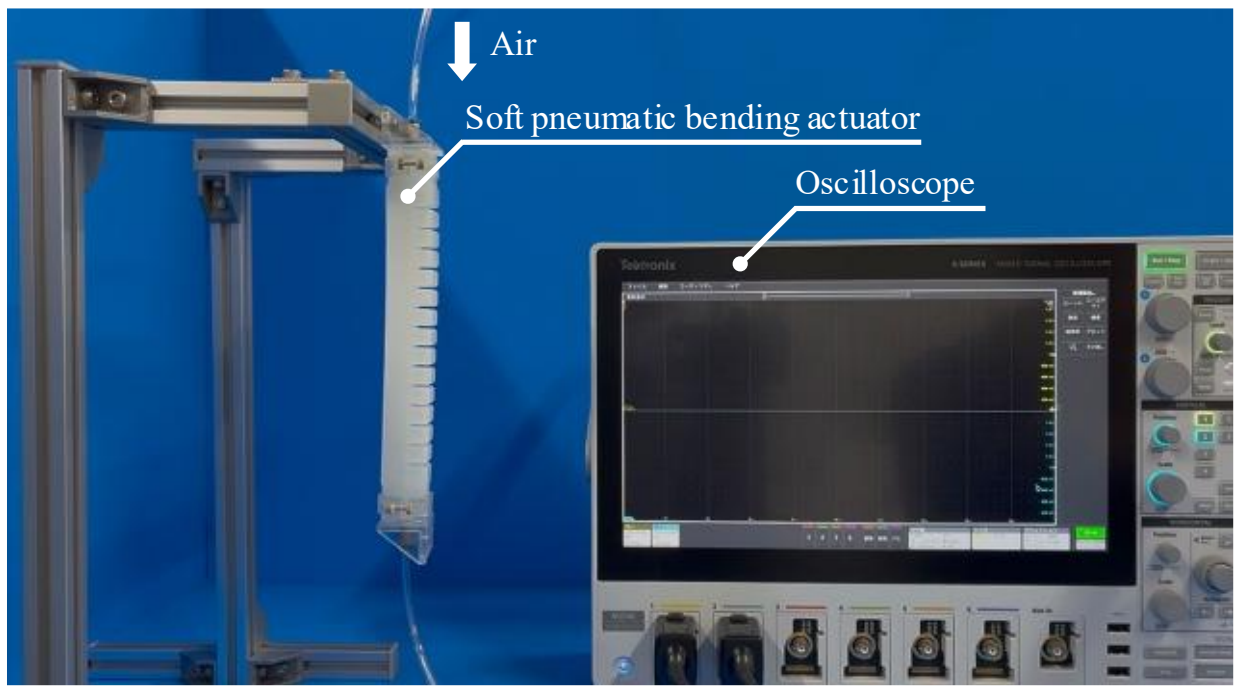

Fig. 7. Experimental setup for measuring the pressure propagation delay.

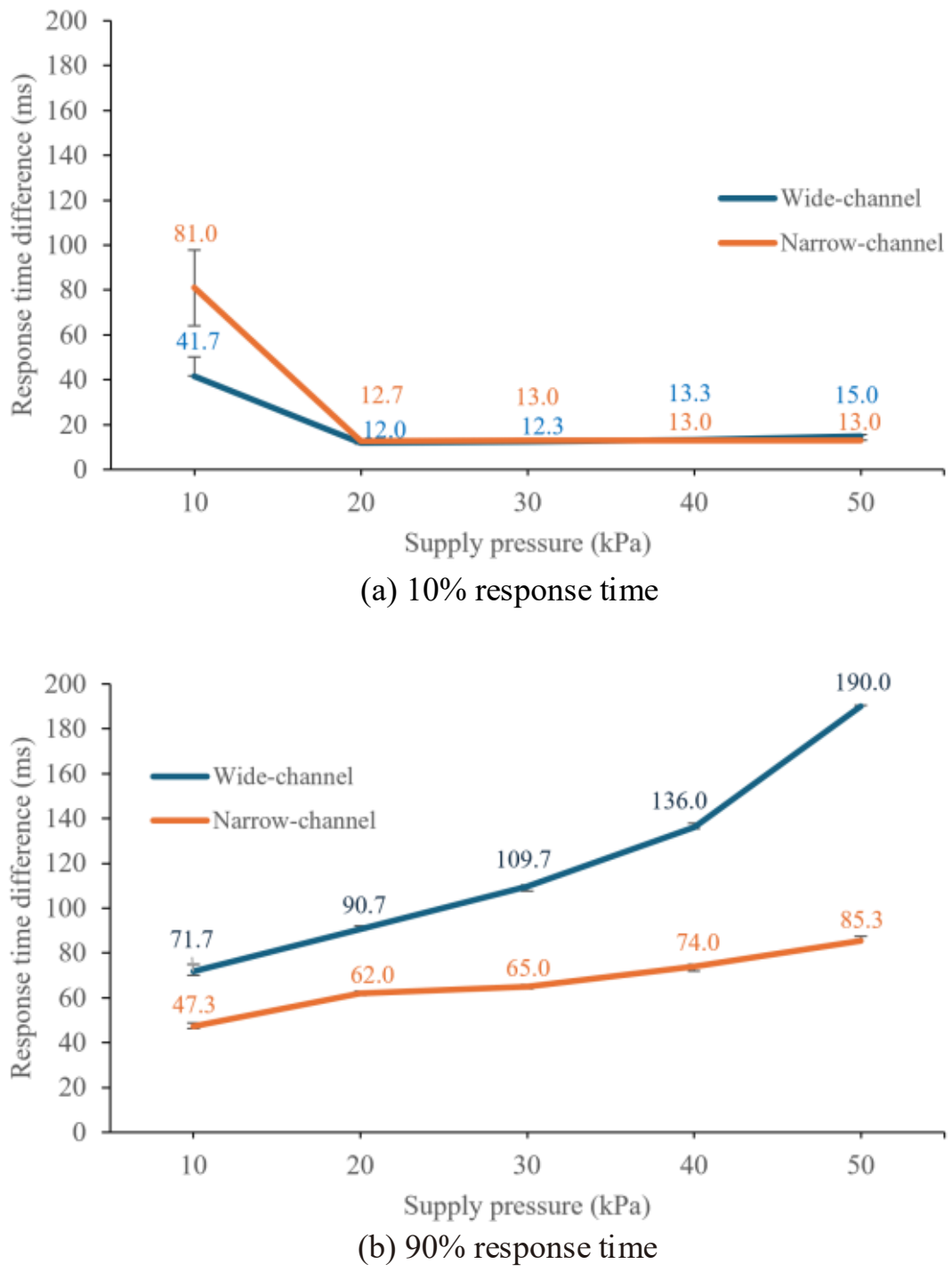

Fig. 8. Response time difference between the proximal and distal ends versus supply pressure for the wide-channel and narrow-channel actuators: (a) 10% response time and (b) 90% response time (mean ± SD, n = 3).

### *C. Discussion and Actuator Selection*

The results in Sections III-A and III-B show that reducing the flow-channel cross-sectional area increases the force required for bending but does not increase the pressure propagation delay.

Because the force was measured without pressurization, the difference in the force required for bending originates from the actuator structure. Narrowing the channel increased the volume of silicone rubber and hence the bending stiffness, so the narrow-channel actuator required approximately 0.12–0.23 N more force.

The pressure propagation delay showed different trends for the 10% and 90% response times. For the 10% response time, the response time difference of the narrow-channel actuator was larger only at a supply pressure of 10 kPa, with almost no difference between the two actuators at 20 kPa or higher. This is likely because the flow restriction of the narrow channel became significant only at low pressures, where the flow rate is small. In contrast, for the 90% response time, the response time difference of the wide-channel actuator was larger and increased with supply pressure. Because the wide-channel actuator has thinner chamber walls and expands more easily, its volume change during pressurization is likely larger, so more air was required for the distal end to reach the steady-state pressure. These results suggest that the delay in the wide-channel actuator arises from expansion rather than from flow restriction in the channel.

In the antagonistic arrangement, the unpressurized actuator resists joint rotation; therefore, a small force required for bending is desirable. Accordingly, the wide-channel actuator, which requires less force for bending, is used in the evaluation of the joint mechanism in Section IV.

## IV. Joint Mechanism Characterization

This section evaluates the characteristics of the prototype as a joint mechanism, using the wide-channel actuator selected in Section III. Section IV-A measures the range of motion, and Section IV-B measures the joint torque. Section IV-C discusses the functional limitations of the mechanism and their causes based on these results, and presents design guidelines for improvement.

In the following experiments, the prototype was installed with the plane of joint rotation horizontal to minimize the effect of gravity on the joint motion. The state in which the two actuators were balanced at $\theta = 0°$ was defined as the initial position.

### A. Range of Motion

Because the Hillberry joint mechanism can achieve large rotation angles, a wide range of motion is also required of the prototype. Therefore, the joint angle of the prototype was measured with a motion capture system (V120:Trio, OptiTrack) to evaluate its range of motion. The experimental setup is shown in Fig. 9. Markers were attached to the links, and 60 kPa was applied to one actuator from the initial position to rotate the joint in the positive $\theta$ direction. This pressure is the upper limit that the fabricated actuators can withstand without damage. Subsequently, the pressurized actuator was vented to the atmosphere, and 60 kPa was applied to the antagonistic actuator to rotate the joint in the negative $\theta$ direction. The measurement was performed three times.

The range of motion of the prototype was limited to −92.5° to 49.0°, which did not fully exploit the wide range of motion that is an advantage of rolling contact joints (Figs. 10 and 11).

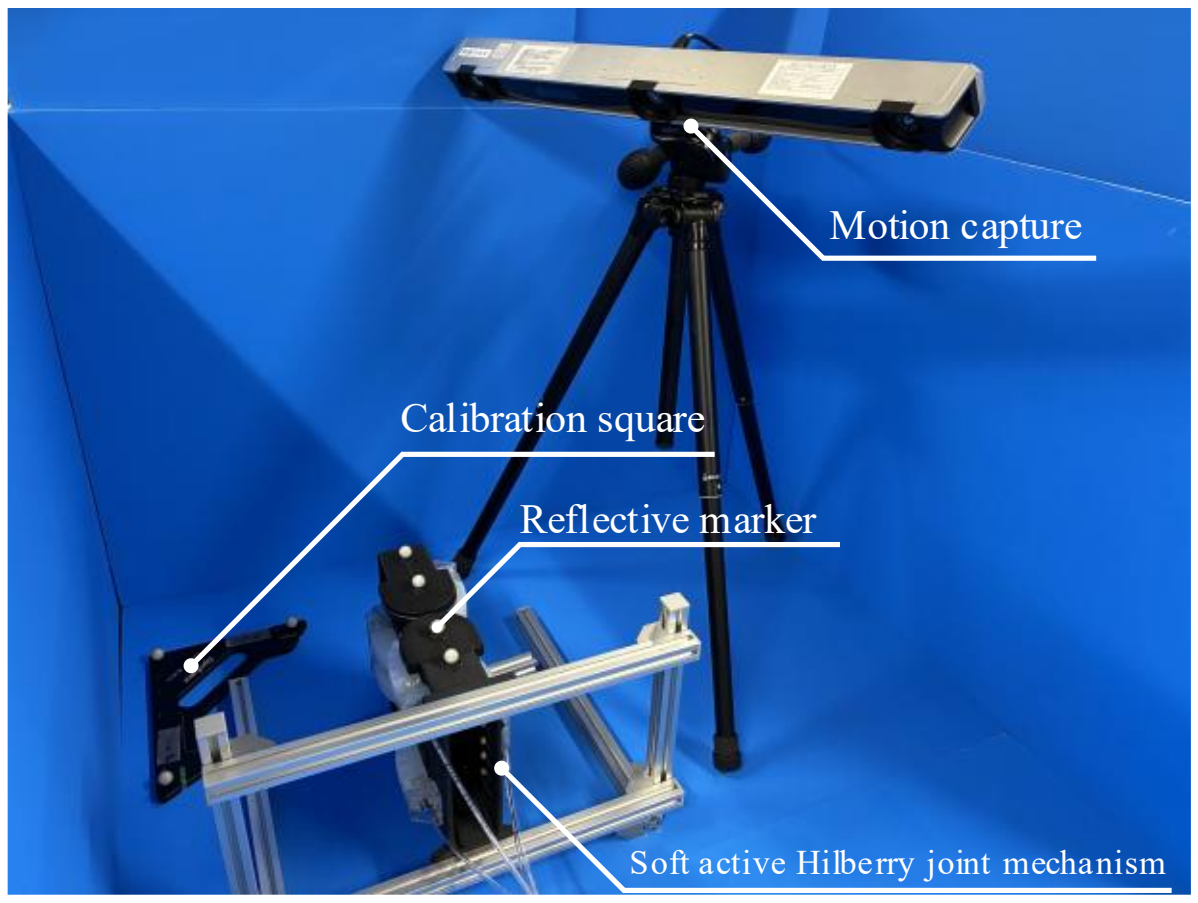

Fig. 9. Experimental setup for measuring the range of motion. The joint angle $\theta$ was measured from the positions of markers attached to the links using a motion capture system.

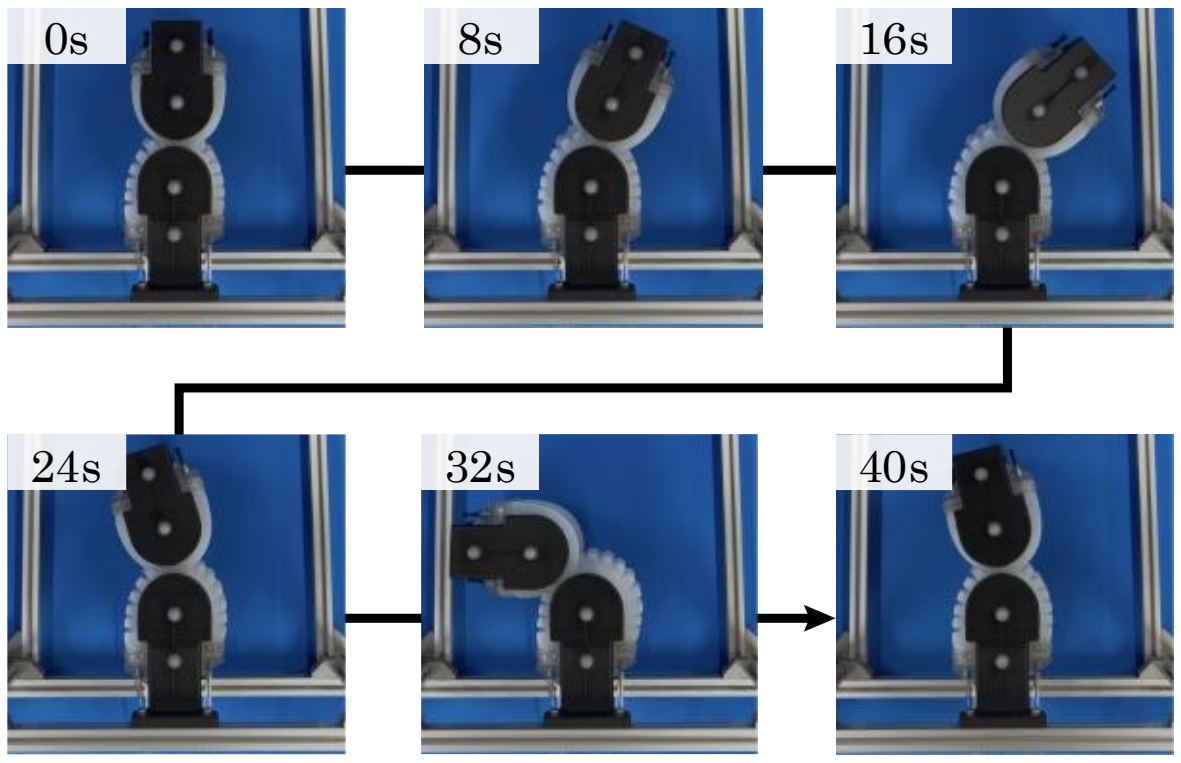

Fig. 10. Sequential snapshots of the prototype with 60 kPa applied first to one actuator and then to the antagonistic actuator.

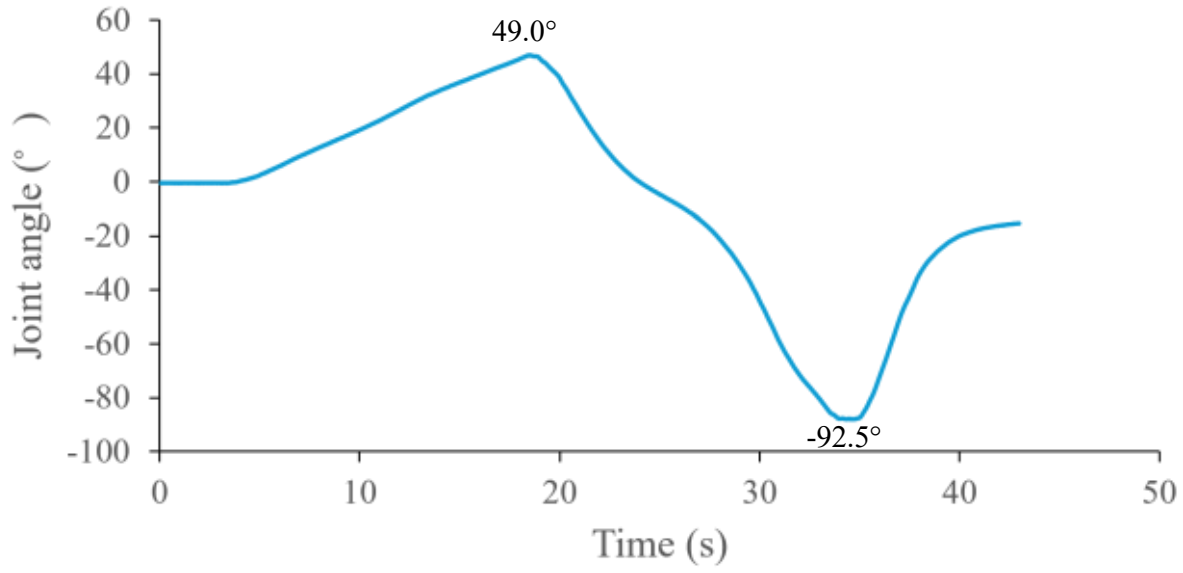

Fig. 11. Time history of the joint angle $\theta$ during the range-of-motion measurement (mean of three trials). $\theta$ increased when one actuator was pressurized to 60 kPa and decreased when the antagonistic actuator was pressurized after the first was vented.

The maximum and minimum values of $\theta$ were 49.0 ± 2.3° and −92.5 ± 2.0°, respectively (mean ± SD, n = 3). The fact that the range of motion remained this limited even at the maximum allowable pressure indicates that the current prototype lacks sufficient driving torque to rotate the joint over a wide range. In addition, the range of motion in the negative direction was approximately twice that in the positive direction, showing an asymmetry between the two directions. This asymmetry is likely due to individual differences between the actuators, as the actuators on the two sides inflated differently during pressurization and θ settled at approximately −13° instead of returning to the initial position at the end of the experiment.

### *B. Joint Torque*

To evaluate the practicality of a joint mechanism, its output torque must be quantified in addition to its range of motion. This evaluation is particularly important for joint mechanisms driven by soft actuators, whose output tends to be limited in exchange for their compliance. Therefore, the joint torque of the prototype was measured using a torque sensor (UTMIII-0.5Nm, UNIPULSE) and an oscilloscope (MSO24, Tektronix). The experimental setup is shown in Fig. 12. In the proposed mechanism, the amount of actuator bending and the state of rolling contact between the links change with the joint angle; therefore, the output torque is expected to depend on the joint angle. To obtain the relationship between joint angle and torque continuously, the joint was externally rotated at a constant speed of 3 rpm by a servomotor (XH430-V210-R, ROBOTIS) while a constant pressure was applied to only one actuator, and the resulting joint torque was measured. The joint was first rotated from θ = 0° to approximately +110° in the direction in which the pressurized actuator rotates the joint (the positive θ direction), then to approximately −110°, and finally back to θ = 0°. The supply pressure was varied from 10 to 50 kPa in steps of 10 kPa, and three measurements were performed for each pressure.

The joint torque exhibited hysteresis at all supply pressures, and the hysteresis width increased with supply pressure (Fig. 13). The hysteresis width averaged over θ = −100° to 100° was 0.19 N·m at 10 kPa and 0.30 N·m at 50 kPa. This increase was mainly due to the torque in the interval in which θ was decreased, which became more negative as the supply pressure increased. This is because decreasing θ requires rotating the joint against the torque generated by the pressurized actuator.

### *C. Discussion and Design Guidelines*

The results in Sections IV-A and IV-B suggest that the range of motion and joint torque of the prototype are limited by two main factors: the deformation resistance of the antagonistic actuator and the wires constraining the links. In the proposed mechanism, regardless of the direction of joint rotation, the antagonistic actuator remains unpressurized and deforms such that its bottom surface is stretched, and the force required for this deformation resists the rotation. As shown in Section III-A, bending an unpressurized actuator required a force of up to approximately 1.0 N, and this resistance offsets part of the torque generated by the pressurized actuator. In addition, the wires attached to maintain the distance between the links restricted actuator expansion.

Based on these findings, two design guidelines are proposed for joint mechanisms driven by soft pneumatic bending actuators. First, the deformation resistance of the antagonistic actuator should be reduced. For example, small slits can also be added to the bottom side of the actuator so that the bottom surface stretches more easily. Another approach is to use a soft actuator that has two separate channels and can bend in both directions [24]. With such an actuator, a single actuator can generate torque in both directions, which may eliminate the resistance arising from the antagonistic arrangement itself. Second, the method of constraining the links should be reconsidered. Without constraints such as wires, actuator expansion changes the distance between the links, and the distal end of the link deflects under gravity. However, the current wires restrict actuator expansion. This problem can be addressed by imparting anisotropic deformation to the actuators, for example through fiber reinforcement, so that the actuators themselves suppress extension and expansion in directions other than bending [25], [26]. This may enable a configuration that maintains the distance between the links without restricting actuator bending.

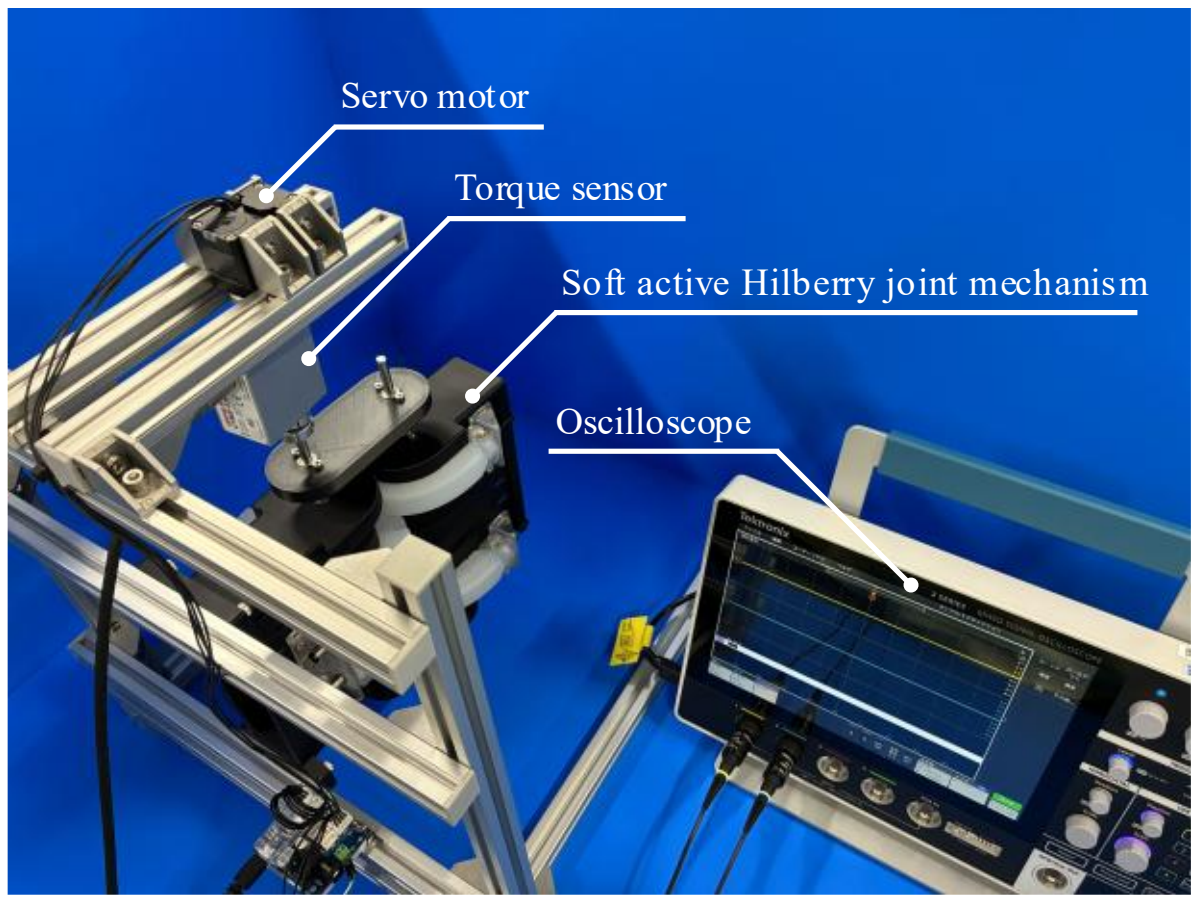

Fig. 12. Experimental setup for measuring the joint torque. The joint was rotated at a constant speed of 3 rpm by a servomotor while one actuator was pressurized, and the torque was measured with a torque sensor.

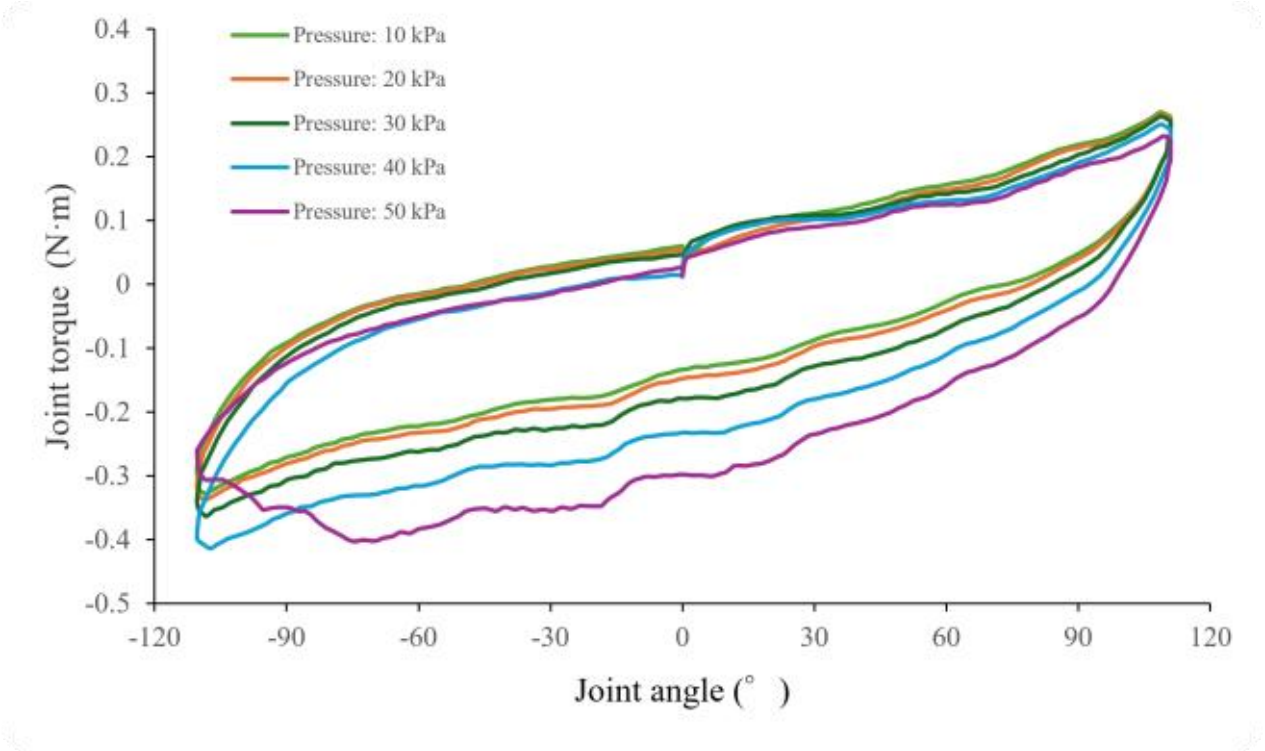

Fig. 13. Joint torque versus joint angle θ for supply pressures of 10–50 kPa applied to one actuator (mean of three trials). The joint was rotated by the servomotor from 0° to approximately +110° (①), then to approximately −110° (②), and back to 0° (③). Positive torque corresponds to the positive θ direction.

## V. Conclusion

This paper proposed a soft active Hillberry joint mechanism in which the rolling contact section is directly driven by antagonistically arranged soft pneumatic bending actuators, and demonstrated its principle with a prototype. A comparison of flow-channel cross-sectional areas showed that narrowing the channel increased the force required for bending by approximately 0.1–0.2 N and did not produce the intended pressure propagation delay. The range of motion of the prototype was limited to −92.5° to 49.0°, and the results suggested that the deformation resistance of the antagonistic actuator and the wires constraining the links are the main limiting factors. Based on these findings, reducing the deformation resistance of the antagonistic actuator and imparting anisotropic deformation to the actuators were presented as design guidelines. Future work will improve the mechanism based on these guidelines and address joint angle control.

## Acknowledgment

The authors used ChatGPT-6 (OpenAI) and Claude Opus 5.5 (Anthropic) to assist with drafting, editing, and translating parts of the manuscript into English, and with discussing the experimental methods. The authors reviewed and revised all AI-assisted content and take full responsibility for the content of this paper.